\documentclass{article}

\usepackage[square,sort,comma,numbers]{natbib}
\usepackage{wrapfig}

\usepackage[final]{nips_2017}

\usepackage[utf8]{inputenc} 
\usepackage[T1]{fontenc}    
\usepackage{hyperref}       
\usepackage{url}            
\usepackage{booktabs}       
\usepackage{amsfonts}       
\usepackage{nicefrac}       
\usepackage{microtype}      
\usepackage{graphicx}
\usepackage{algorithm}
\usepackage[noend]{algpseudocode}

\title{Achieving Fluency and Coherency in Task-oriented Dialog}

\author{
  Rashmi Gangadharaiah, Balakrishnan (Murali) Narayanaswamy, Charles Elkan \\
  Amazon AI Lab \\
 \{\texttt{rgangad@,muralibn@,elkanc@\}amazon.com}\\
}

\begin{document}

\maketitle

\begin{abstract}
We consider real world task-oriented dialog settings, where agents need to generate both fluent natural language responses and correct external actions like database queries and updates. We demonstrate that, when applied to customer support chat transcripts, Sequence to Sequence (Seq2Seq) models often generate short, incoherent and ungrammatical natural language responses that are dominated by words that occur with high frequency in the training data. These phenomena do not arise in synthetic datasets such as bAbI, where we show Seq2Seq models are nearly perfect. We develop techniques to learn embeddings that succinctly capture relevant information from the dialog history, and demonstrate that nearest neighbor based approaches in this learned neural embedding space generate more fluent responses. However, we see that these methods are not able to accurately predict when to execute an external action. 
We show how to combine nearest neighbor and Seq2Seq methods in a hybrid model, where nearest neighbor is used to generate fluent responses and Seq2Seq type models ensure dialog coherency and generate accurate external actions. 
We show that this approach is well suited for customer support scenarios, where agents' responses are typically script-driven, and correct external actions are critically important. The hybrid model on the customer support data achieves a $78\%$ relative improvement in fluency scores, and a $130\%$ improvement in accuracy of external calls.
\end{abstract}

\section{Introduction}
\label{sec:intro}

Large or open domain chatbots are now omni-present, reaching many people through services like Microsoft's Cortana, Google's Assistant, Amazon's Alexa or Apple's Siri \cite{dale2016return}. Recent efforts have been more focussed towards basic chit-chat \cite{vinyals:15, shang:15, serban:15, sordoni:15, dodge:16} that are non-goal oriented. Chit-chat here refers to the ability to generate fluent responses, that are reasonable in the context of the conversation. In contrast, in task or goal oriented dialog, the chatbot needs to extract relevant information from the user (e.g. preferences), provide relevant knowledge to her (e.g. prices and availability), and issue appropriate system calls (e.g. make a payment).  Very few approaches have been applied to goal-oriented settings but evaluated on synthetic datasets, such as bAbI \cite{bordes:17} or through Wizard Of Oz experiments. 

Advances in training deep neural networks has demonstrated the potential to build chatbots with minimal expert domain knowledge. In particular, supervised approaches such as Seq2Seq learning \cite{vinyals:15}, can perform end-to-end learning from expert trajectories or dialogs, removing the need for many of the independent modules in traditional dialog systems, such as, the natural language understanding component, the natural language generation component, the dialog policy and the state tracker.
Seq2Seq models (Section \ref{sec:basic}) achieve almost perfect performance on such synthetic datasets, however, they generate short, non-fluent and incoherent responses when tested on a real world dataset. 

\textbf{Fluency and Coherency}: An example dialog generated by our best Seq2Seq model, trained on real customer-agent chat transcripts is shown in Table \ref{gen-table}. We say a response is \textit{coherent} if it is a sensible or a logical response considering the dialog context. Note that a coherent response does not necessarily guarantee fluency or grammaticality. For example, a sensible, coherent response at turn 4 in Table \ref{gen-table} would be to accept the customer's expression of gratitude as opposed to saying \textit{how can I help you today ?} Consider the generated response at turn 2, the response is coherent because the agent is thanking the customer for being a member which is a sensible response that could occur at the beginning of the conversation, even though the response is ungrammatical or not fluent. Ideally, we want responses to be both \textit{coherent} as well as \textit{fluent} to ensure that the intended information is both conveyed and understood.

Coherence is a dialog level abstraction, any measure of which should correlate with how appropriate a response is in the context of the entire dialog history, while fluency is a measure of grammatical correctness of agents' responses at an utterance level. Measuring coherency is difficult in general. In our task oriented problems, we use task success as a measure of coherency. However, for example, consider the response at turn 5, the generated response is both incoherent as well as inarticulate, and fails to convey or represent the right information. Measuring dialog level coherence is important, and something we will address in future work. 

We see that, when applied to real dialog datasets, Seq2Seq based models perform well with salutations and external actions but perform poorly on intermediate responses, a phenomenon we see throughout the dataset. In particular, they generate responses that are short, incoherent and ungrammatical. For example, the average utterance length generated by Seq2Seq is 4.2 words, compared with ground truth lengths that average 11.6. This is in part because end to end methods in general and Seq2Seq based models in particular, require large amounts of data before they are able to generate even somewhat fluent textual responses. To mitigate some of these problems, we explore nearest neighbor-based approaches. We show that such an approach is well suited to customer support scenarios, since the agents' responses are typically script-driven.


\begin{table*}[t]
  \caption{Generated dialog on customer support data. \textit{masked} used for anonymization.}
  \label{gen-table}
  \tiny
  \centering
  \begin{tabular}{llll}
    \toprule
   Turns &  User's  Utterance    & Agent's Response   & Generated Agent's Response\\
    \midrule
   Turn 1 &  got charged for <masked> membership & hello , my name is <PERSON> .  i 'm here to &  hello <PERSON> ,  my name is <PERSON> .\\
   & after trial i did not want to continue.  & help you today . &   i 'm here to help you \\ \hline
   Turn 2  & <SILENCE> & thank you for being a <masked> member . & thank , for you <masked> member \\ \hline
  Turn 3  & <SILENCE>  & i 'm sorry to hear that you were charged with & <SILENCE>   \\
  & &  our membership . no worries i 'll do my best & \\ \hline
   Turn 4 & thank you , i did not know that it  & you 're welcome & you 're most welcome \\
   &  would auto renew after the trial & & \\ \hline
   Turn 5 & <SILENCE> & i will now cancel your membership & i you sorry you while and you will issued \\ \hline
   Turn 6 & <SILENCE> & $api\_call$ cancel\_refund & $api\_call$ cancel\_refund \\ \hline
   Turn 7 & thank you so much & i hope my actions helped you out today & you 're most welcome \\ \hline
   Turn 8 &<SILENCE> & you 're most welcome & is there anything else i can assist you with with \\ \hline
   Turn 9  & very appreciated & is there anything else i can help you with today ? & <SILENCE> \\ \hline
   Turn 10 & no i 'm good & thank you for contacting <masked> &  thank you for contacting <masked> \\
    \bottomrule
  \end{tabular}
\end{table*}

\section{Related Work}
\label{sec:relwork}

Our work is closely related to retrieval-based chatbots \cite{ji:2014, zhou:2016, Yan:2016}. These approaches formulate queries by concatenating all the utterances in the dialog history to create a bag of words representation to retrieve a response from a known set of responses. These approaches evaluate precision@K, from a restricted list, but do not indicate how this restricted list is obtained in practice. In contrast, we use a learned fixed size vector representation as a succinct summary of the dialog history, borrowing ideas from Seq2Seq generation-based chatbots \cite{vinyals:15, serban:15, shang:15}. 

Multi-class classification-based approaches \cite{bordes:17} also pick one response (or class) from a fixed set of responses. A big disadvantage with such an approach is generating negative examples for each class. These approaches use poor selection strategies such as random sampling from the dialogs in the training data. This approach performed poorly when tested on our dataset, with most of the predicted responses being salutations, since salutations occur much more often in real world dialogs than rare problems. As the number of possible responses can grow significantly over time due to the heavy tailed nature of customer problems, such an approach increases the complexity of the dialog systems. In contrast, we explore nearest neighbor approaches that use vector representations or embeddings obtained from the Seq2Seq-based models, and show that the performance scales well with the number and variety of dialogs.

Since many of the previously proposed approaches to task oriented dialog have been evaluated on synthetic datasets, it has been unclear how these approaches perform on real world tasks. In this paper, we propose an approach to generate both fluent and coherent responses, and test the approach on an internal customer support dataset.


\section{Proposed Approach}
This section describes our nearest neighbor approach for response selection. We see that such an approach is not able to accurately predict when to execute an external action. We show how to combine the nearest neighbor approach and Seq2Seq methods in a hybrid model, where nearest neighbor is used to produce fluent responses and Seq2Seq type models ensure dialog coherency and generate accurate external actions. 
We will explain each in some detail, with experiments showing the benefits of each and all together in combination. First we describe the datasets and metrics we use in our evaluations.

\subsection{Dataset and Metrics}
\label{sec:met}
We first use data from the bAbI Tasks (Task1 and Task2) to evaluate our models. The other dialog tasks in bAbI require the model to mimic the knowledge base i.e., learn the entries in the knowledge base, making the knowledge base less useful once the model is trained. This is not a suitable strategy for our application, since knowledge bases undergo very frequent changes. In the bAbI task, the user interacts with an agent, in a simulated restaurant reservation application, by providing her constraints, such as place, cuisine, number of people or price range. The agent or chatbot performs external actions or SQL-like queries ($api\_call$) to retrieve information from the knowledge base of restaurants, and make reservations. We refer to an entire session of text exchanges between an agent and a customer as a \textit{dialog} and a \textit{turn} refers to one interaction or a pair of text exchanges between the agent and the customer. Note that a system built using hand made rules can achieve $100\%$ accuracy on this dataset, since it is synthetically created using a rule based generator. However, the goal of this paper and that dataset is not just to measure absolute performance, but to identify shortcomings of Seq2Seq-based models in a task oriented setting with no domain specific information. In our experiments, 80\% of the data was used for training (of which 10\% was used for validation) and the remaining 20\% for testing the models. All the models were evaluated on the same test set. 

In addition to the bAbI dataset, we also evaluate our models on an internal customer support dataset. An example dialog from this dataset is shown in Table \ref{gen-table}. A customer support agent in this application, receives issues that belong to one of $31$ possible categories. We consider one category, a subset of account issues, where solutions provided by agents were limited. We randomly sampled 1000 chat transcripts where the dialogs were not escalated, and also limited the number of turns to 20. Ideally we would want the model to also respond when user deviates into tangential topics, however, this would require the chatbot to know and understand general topics, but that is not the focus here. 

The vocabulary size, maximum number of turns per dialog and the number of words per turn are much bigger than the bAbI dataset, making the task  challenging. In addition, these transcripts are very noisy, with typographical errors. We perform spell correction, de-identification to remove customer sensitive information (such as names and addresses), lexical normalization particularly of lingo words such as \textit{lol} and \textit{ty}. Generalizing such entities reduces the amount of training data required. The values can be reinserted \cite{williams:16, wen:17} into the generated response. To preserve anonymity, we mask product and the organization name in the examples. One caveat to our evaluations is that they are based on customer responses to the actual human agent interactions, which are not fully indicative of how customers would react to the real automated system in practice. This is the difficulty of off-policy evaluation and learning. In ongoing work, we evaluate the system with real human agents providing utterance level scores for every dialog.

The fluency of dialog system outputs is typically evaluated using BLEU (BiLingual Evaluation Understudy \cite{papineni:02}), which has been shown to highly correlate with human evaluation, BLEU is piecewise constant, and so models are usually trained to optimize categorical cross entropy loss \cite{lowe:15, liu:16}. An advantage of high BLEU scores is that they indicate that the chatbot would produce similar utterances in similar situations - reducing the problem of off-policy evaluation mentioned above. 
It should be noted that computing BLEU at turn level is also impacted by coherency since the generated response is compared against the response provided by a human agent. The use of evaluation metrics like BLEU or METEOR \cite{lavie:07}, to evaluate dialogs with just one reference has been debated \cite{liu:16}. 
There is still no good alternative to evaluate dialog systems, and so we continue to measure fluency using BLEU here. 

Coherency also requires measuring correctness of the external actions which we measure using a metric we call Exact Query Match (EQM), which represents the fraction of times the $api\_call $ matched the ground truth query and we do not assign credits to partial matches. In addition, we report the precision (P), recall (R) and accuracy (Acc) achieved by the models in predicting whether to make an $api\_call$ (positive) or not (negative). These metrics help measure the timing accuracy of the api calls.

\begin{table*}[t]
  \caption{Results with variants of the Seq2Seq model on the bAbI dataset.
 }
  \label{Seq2SeqMResults}
  \small
  \centering
  \begin{tabular}{lllllllll}
    \toprule
   Model & Type &  Description &  BLEU & P & R & Acc & EQM\\
    \midrule
   Model 1 & Basic Seq2Seq &  dependencies between turns absent &  88.3 &  0.60 & 1.00 & 0.87& 0.00\\
  Model 2  & HRED Seq2Seq & Model1 + append $h^{t-1}_{L,enc}$ & 90.2 &  1.00 & 1.00 & 1.00 & 0.06 \\
    \bottomrule
  \end{tabular}
\end{table*}

\subsubsection{Seq2Seq Model}
\label{sec:basic}
\begin{figure*}[h]
    \centering
    \includegraphics[width=0.8\textwidth]{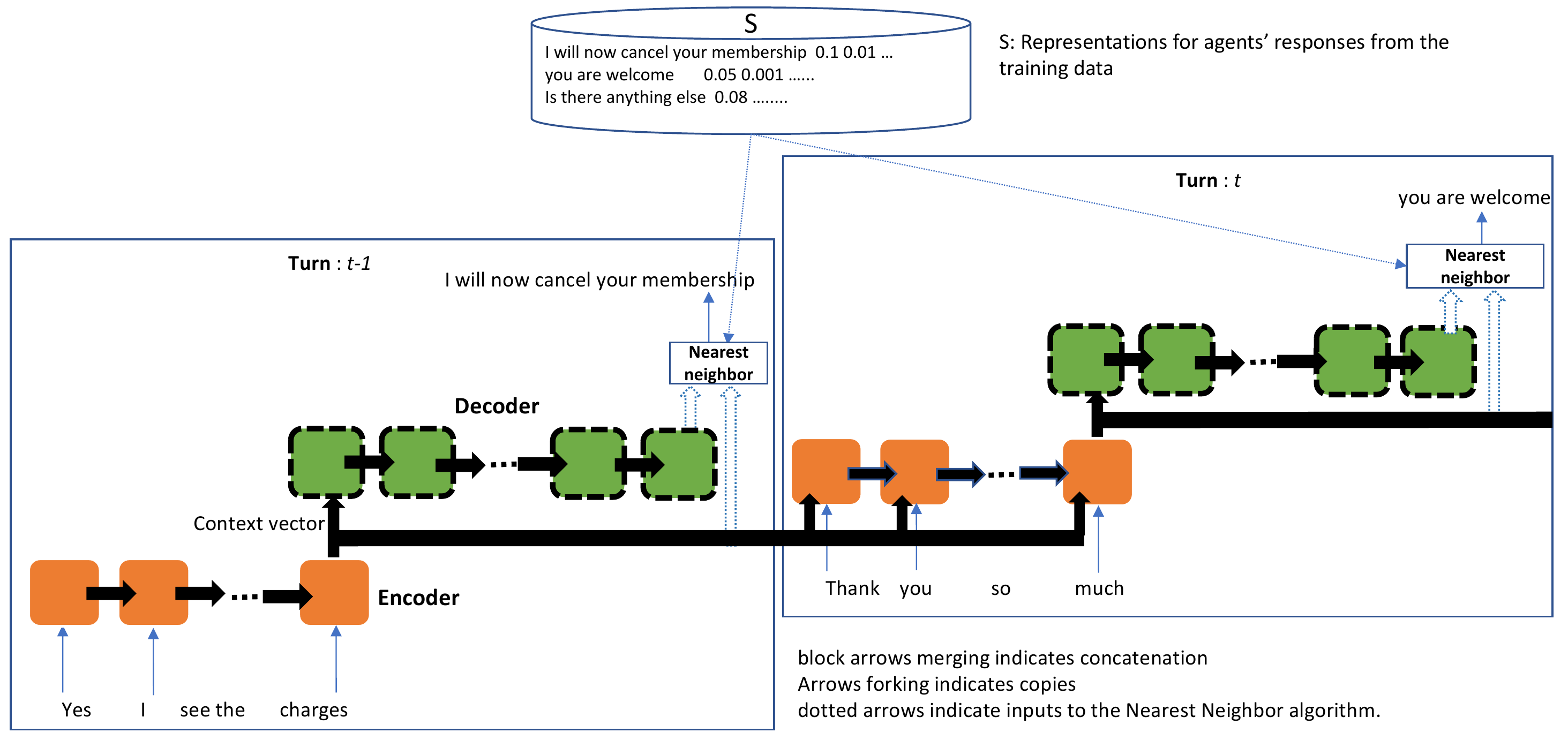}
    \caption{Proposed embeddings for finding the nearest neighbor.}
    \label{fig:NNB_Seq2Seq}
\end{figure*}

The basic Seq2Seq model \cite{sutskever:14} consists of two components, an encoder and a decoder, typically modeled using Long Short Term Memory Units (LSTMs) \cite{Hochreiter:1997}. The encoder encodes the input sequence into a fixed dimensional vector, the output of the last hidden layer of the encoder. This vector is input to the decoder, which generates the output. In customer support, the input sequence is the user's utterance and the output is the agent's response. We use a novel variant of the hierarchical recurrent encoder-decoder (HRED) \cite{sordoni:15}, to extend the basic Seq2Seq to handle multi-turn dialog. We unroll the basic Seq2Seq model, and make one copy for each turn present in the dialog. In Figure \ref{fig:NNB_Seq2Seq}, orange-solid-square boxes represent the embedding and LSTM cells of the encoder. Green-dashed-square cells in the decoder represent the LSTM and dense layers with softmax for predicting each word in the agent's utterance or query sequence. The block arrows represent flow of information from one cell to another and arrows forking represent copies of this information. Each of the Seq2Seq copies share the same parameters. Once the training is complete, we use only one copy of the Seq2Seq model to make predictions. In order to make predictions for agent's responses at turn \textit{t}, the context vector at turn \textit{t-1} is appended to the embeddings of the user's utterance in turn \textit{t}. 

Table \ref{Seq2SeqMResults} shows results obtained with the vanilla Seq2Seq model which does not handle dialog history or context and results obtained with HRED. It can be seen that adding dependencies between turns substantially improved the perfomance across all the measures. 

An improvement was also seen on the internal customer support dataset with HRED (Model 2 in Table \ref{NNResults}). However, the generated responses were most often incoherent and not fluent. Table \ref{gen-table} shows an example response generated by a HRED variant, Model 2. Comparing the generated agent's response (column 4) with the true response (columns 3), it is evident that the model performs poorly on intermediate responses. This is because the outputs of Seq2Seq are dominated by words that occur with high frequency in the training data. As an example, $you$ occurs with probability $>0.1$ in the output of Seq2Seq, which occurs about one tenth as frequently in the training data. At a bigram level, much of the probability mass is assigned to ungrammatical combinations of frequent unigrams like \textit{i you}. 

We now proceed to explain the nearest neighbor based approach to produce reasonable responses that are more fluent.
\subsection{Nearest Neighbor-based approach}
\label{sec:NewModel}
In our nearest neighbor based approach to generating utterances, an agent's response is chosen from human generated transcripts - the training data - ensuring fluency. However, this does not necessarily ensure that the responses are coherent. The nearest neighbor approach starts with a belief state vector or embedding obtained from the entire history of the dialog so far, to improve coherency. 
\begin{algorithm}[H]
\caption{Nearest Neighbor-based (NNB) approach}
\label{NNB}
 \scriptsize
\begin{algorithmic}
\item \textbf{for} $i=1:D$, \Comment{$D=$ number of test dialogs}
\item \ \ \ \textbf{for} $t=1:T$, \Comment{$T=$ max no. of turns in the test dialog$_i$}
\item \ \ \ \ \ \ \textbf{if} $t==1$
\item \ \ \ \ \ \ \ \ \ $testVec=$BeliefStateRepresentation($user_t$)
\Comment{$user_t$ is the response of the user at turn $t$,}
\item \ \ \ \ \ \ \textbf{else}
\item \ \ \ \ \ \ \ \ \ $testVec=$BeliefStateRepresentation($user_{1:t}$, $agent_{1:t-1}$)
\Comment{$agent_{p}$ is the $true$ response of the agent at turn $p$}\\
\item \ \ \ \ \ \ \ $bs_{testVec}=$NNB ($testVec$, $S$) \Comment{find nearest neighbor of $testVec$ in $S$}
\item \ \ \ \ \ \ \ \textbf{return} $a_{testVec}$, agent's response represented by $bs_{testVec}$
\end{algorithmic}
\end{algorithm}

\begin{table*}[t]
  \caption{Results with the Nearest Neighbor approach on customer support data. 
 }
  \label{NNResults}
  \small
  \centering
  \begin{tabular}{lllllllll}
    \toprule
   Model & Description &  BLEU & P & R & Acc & EQM\\
    \midrule
   Model 2 & Seq2Seq & 9.91  & 0.34  & 0.79  & 0.81 & 0.30\\
   Model 3 & Nearest neighbor using output of encoder & 15.14 & 0.38 & 0.35 & 0.86 & 0.13 \\
   Model 4 & Nearest neighbor using output of decoder & 16.34 &  0.36 & 0.31 & 0.86 & 0.16 \\
   Model 5 & Best Of both (Models 2+4) & 17.67 & 0.33 & 0.73 & 0.80 & 0.30\\
    \bottomrule
  \end{tabular}
\end{table*}

\begin{table*}[t]
  \caption{Generated dialog on customer support data using the hybrid model. \textit{masked} used for anonymization.}
  \label{result-table}
  \tiny
  \centering
  \begin{tabular}{llll}
    \toprule
   \textbf{Turns} &  \textbf{User's  Utterance}    & \textbf{Agent's Response}   & \textbf{Generated Agent's Response} \\
    \midrule
   Turn 1 &  i want to cancel my <masked> 
 & hello <PERSON> , my name is <PERSON> .  & hello , my name is <PERSON> . \\
& membership , please refund me & thank you for being a <masked> member . & i 'm here to to help you today .\\ \hline
   Turn 2  & hi , i found a bill list that charged & <SILENCE> & please allow me a minute to check this for you . \\ 
   & me <MONEY>& &\\ \hline
  Turn 3  & i did not order <masked> membership  & <SILENCE>  & thanks for the details . \\ \hline
   Turn 4 & can you help me & please do not worry , i 'll be completely  & i 'm sorry if any inconvenience happened to you .\\
 & & helping you with this . & \\ \hline
   Turn 5 & thank you very much & you 're welcome . & i have successfully issued the refund for you and\\
   & & & i will make sure this does not happen in future \\
   & & & again .\\ \hline
   Turn 6 & <SILENCE> & <masked> successfully canceled , <MONEY> & $api\_call$ cancel\_refund \\
   & & refund processes .  &\\ \hline
   Turn 7 & <SILENCE> & $api\_call$ cancel\_refund & $api\_call$ cancel\_refund \\ \hline
   Turn 8 & okay , good job & i 've already processes the refund in . & it was my pleasure assisting you today . thank \\
   & & & you for contacting <masked> and do have a \\
   & & & lovely time .\\ \hline
   Turn 9 &<SILENCE> & i 'm glad i could help you with this . & thank you for contacting <masked> , again this\\
   & & & is <PERSON> and it 's my pleasure assisting you \\
   & & & today .\\ \hline
   Turn 10  & <SILENCE> & is there anything else i can help you with ?
 & in the meantime , i want to make sure i have \\
 & & & covered all of your concerns , please let me know \\ \hline
   Turn 11 & no more , thanks
 & it was my pleasure assisting a valued 
 &  please click on the end chat to close this window\\
 & & customer like you today . &\\
    \bottomrule
  \end{tabular}
\end{table*}

\subsection{Nearest neighbor with Seq2Seq embeddings}
\label{sec:ours}
We propose using the outputs of the hidden units in the decoder and the encoder as suitable representations of the belief states. The Seq2Seq model for handling multi-turn dialog is trained as before (Section \ref{sec:basic}). Once the parameters have been learned, we proceed to generate representations of the state before each agents' responses in the training data. Pseudo-code for the nearest neighbor approach is in Algorithm \ref{NNB}. 
As seen in Section \ref{sec:basic}, the output of the last hidden unit of the encoder or the decoder at turn $t$, capture the dialog history well. This results in a tuple $<bs_{t,i},a_{t,i}>$, where, $bs_{t,i}$ represents the belief state at turn $t$ for dialog example $i$ and $a_{t,i}$ represents the action the agent took while in this state i.e., the natural language response or $api\_call$ query issued by the agent. This is done to obtain belief state representations for all actions or agents' responses in the training set. This results in a set $S$ that contains pairwise relationships between belief states and agent's actions. An agent's responses are scripted, and often multiple belief states will have the same action. 

We test the models as done in Section \ref{sec:basic}, except now we do not generate the agent's response directly from the decoder. Figure \ref{fig:NNB_Seq2Seq} summarizes this process. We use the output of the last hidden unit of the decoder, $testVec$, to find the nearest neighbor $bs_{testVec}$ in $S$. We return the nearest neighbor's corresponding response, $a_{testVec}$, as the predicted agent's response. We use ball trees \cite{Kibriya:2007} to perform efficient nearest neighbor search. We can also concatenate the output of the last hidden unit of the encoder to the decoder's output to represent the belief state vector, shown as dotted arrows in Figure \ref{fig:NNB_Seq2Seq}. We experimented with all three combinations, (1) considering the encoder's output, (2) considering the decoder's output (3) concatenating both. 

Results obtained with this approach are in Table \ref{NNResults} (Models 3 and 4). Model 3 uses the output of the last hidden unit of the encoder only. Model 4 uses the output of the last hidden unit of the decoder. Both the models show a significant improvement in BLEU when compared to generating the agent's response directly from the Seq2Seq model (Model 2). 
Appending both the encoder's and the decoder's output did not show a significant change and hence, we do not report the results here. 

The results also show that the Seq2Seq model achieved a better EQM when compared to the nearest neighbor approach (Models 3 and 4). The final hybrid model we propose, combines the best of both the strategies. We run both the Models 2 and 4 in parallel, when Model 2 predicts an API response, we use the output generated by Model 2 as the agent's response, otherwise we use the output of Model 4 as the predicted agent's response. This model achieved the best results among all models we study, both in terms of fluency (BLEU) as well as correctness of external actions (EQM). The hybrid model achieves a $78\%$ relative improvement (from 9.91 to 17.67) in fluency scores, and $130\%$ improvement in EQM over previous approaches (from 0.13 to 0.30).

\section{Conclusions and Future Work}
In this paper, we demonstrated limitations of Seq2Seq approaches, particularly on real world datasets. We proposed approaches, that combined the strengths of Seq2Seq models and nearest neighbor-based methods for task oriented dialog. We showed that this hybrid model was able to produce coherent and fluent responses. Table \ref{result-table} shows an example response of the hybrid model. While many of the responses semantically match the true agent's response, they do not completely match the true response lexically. For example, \textit{in the meantime, i want to make sure i have covered all your concerns, please let me know} matches \textit{is there anything else i can help you with}, however, the choice of words to represent the intention is completely different. A disadvantage of using evaluation strategies such as BLEU, is that they unnecessarily penalize such valid responses. Although certain predicted responses do not match the true response semantically, they are still reasonable responses. Consider the predicted response, \textit{i'm sorry for any inconvenience happened to you}, where the true response is, \textit{please do not worry , i'll be completely helping you with this}, the responses convey different intentions but both are reasonable responses. The nearest neighbor approach produces redundant responses, for example, \textit{$api\_call$ cancel\_refund} is repeated in turn 6 and 7. As future work we will work on finding strategies to prevent such repetitive responses. 

\bibliography{WSDM}
\bibliographystyle{unsrt}
\end{document}